\documentclass[format=sigconf,anonymous=false,review=false]{acmart}

\AtBeginDocument{%
  \providecommand\BibTeX{{%
    \normalfont B\kern-0.5em{\scshape i\kern-0.25em b}\kern-0.8em\TeX}}}

\copyrightyear{2022}
\acmYear{2022}
\setcopyright{acmcopyright}
\acmConference[SIGIR '22] {Proceedings of the 45th International ACM SIGIR Conference on Research and Development in Information Retrieval}{July 11--15, 2022}{Madrid, Spain.}
\acmBooktitle{Proceedings of the 45th International ACM SIGIR Conference on Research and Development in Information Retrieval (SIGIR '22), July 11--15, 2022, Madrid, Spain}
\acmPrice{15.00}
\acmISBN{978-1-4503-8732-3/22/07}
\acmDOI{10.1145/3477495.3531911}

\makeatletter
\def\blfootnote{\gdef\@thefnmark{}\@footnotetext}
\makeatother

\usepackage{graphicx}
\usepackage[caption=false]{subfig}
\usepackage{color}
\usepackage{algorithm}
\usepackage{algorithmic}

\usepackage{microtype}
\usepackage{amsmath,amsfonts}
\usepackage{mathtools}
\usepackage{mathrsfs}
\usepackage{bm}
\usepackage{booktabs, multirow}
\usepackage{makecell}
\usepackage{stfloats}
\usepackage{xcolor}
\usepackage{arydshln}
\usepackage{pifont}
\usepackage{tikz}
\newcommand*{\circled}[1]{\lower.7ex\hbox{\tikz\draw (0pt, 0pt)%
    circle (.5em) node {\makebox[1em][c]{\small #1}};}}

\begin{document}
\fancyhead{}

%% The "title" command has an optional parameter, allowing the author to define a "short title" to be used in page headers.
\title{Posterior Probability Matters: Doubly-Adaptive Calibration for Neural Predictions in Online Advertising}

\author{Penghui Wei, Weimin Zhang, Ruijie Hou, Jinquan Liu, Shaoguo Liu, Liang Wang and Bo Zheng}
\affiliation{
  \institution{Alibaba Group}
  \city{Beijing}
  \country{China}
}
\email{{wph242967,dutan.zwm,ruijie.hrj,jinquan.ljq,shaoguo.lsg,liangbo.wl,bozheng}@alibaba-inc.com}

\begin{abstract}
Predicting user response probabilities is vital for ad ranking and bidding. We hope that predictive models can produce accurate probabilistic predictions that reflect true likelihoods. Calibration techniques aim to post-process model predictions to posterior probabilities. Field-level calibration -- which performs calibration w.r.t. to a specific field value -- is fine-grained and more practical.  In this paper we propose a doubly-adaptive approach \textsf{AdaCalib}. It learns an isotonic function family to calibrate model predictions with the guidance of posterior statistics, and field-adaptive mechanisms are designed to ensure that the posterior is appropriate for the field value to be calibrated. Experiments verify that \textsf{AdaCalib} achieves significant improvement on calibration performance. It has been deployed online and beats previous  approach. 
\end{abstract}

%
% The code below should be generated by the tool at
% http://dl.acm.org/ccs.cfm
% Please copy and paste the code instead of the example below. 
%
\begin{CCSXML}
<ccs2012>
    <concept>
        <concept_id>10002951.10003227.10003447</concept_id>
        <concept_desc>Information systems~Computational advertising</concept_desc>
        <concept_significance>500</concept_significance>
        </concept>
 </ccs2012>
\end{CCSXML}
\ccsdesc[500]{Information systems~Computational advertising}

\keywords{response prediction, field-level calibration, adaptive mechanisms}

%% This command processes the author and affiliation and title information and builds the first part of the formatted document.
\maketitle

\section{Introduction}
In online advertising systems, predicting user response probabilities, such as click-through rate (CTR) and conversion rate (CVR), is the key for ranking candidate ads and bidding strategies. Take  cost-per-conversion (CPA) system as an example, an ad's expected revenue per thousand impressions (eCPM) is computed by $\mathrm{eCPM} = 1000\cdot \hat p_\mathrm{CTR}\cdot \hat p_\mathrm{CVR}\cdot \mathrm{Bid}_\mathrm{CPA}$, where $\mathrm{Bid}_\mathrm{CPA}$ denotes the bid price the advertiser is willing to pay for one conversion. $\hat p_\mathrm{CTR}$/$\hat p_\mathrm{CVR}$ denotes the predicted CTR/CVR of the ad,  produced by predictive models like logistic regression~\cite{he2014practical} and deep neural networks~\cite{covington2016deep,cheng2016wide}. 

In terms of bidding and charging, we expect that predicted probabilities reflect true likelihoods: if the ad is displayed sufficiently many times and its  CTR is 0.02, the ideal situation is that the predicted value is near 0.02. 
Inaccurate predictions (i.e., over- and under-estimation) will hurt user experiences, prevent advertisers from achieving their marketing goals, and affect the revenue of advertising platforms. 
Predictive models are typically trained with binary labels, because for each displayed ad we can only observe whether a response happens while we do not know the likelihood. Thus there is a discrepancy between model predictions and true likelihoods. Existing studies of predictive models focus more on ranking ability~\cite{zhou2018deep,ma2018entire,wei2021autoheri,xu2022ukd} and pay less attention to this. 

To alleviate such discrepancy, \textit{calibration} aims to post-process model predictions to posterior probabilities~\cite{mcmahan2013ad,borisov2018calibration}. 
Performing fine-grained calibration is more practical in real world systems. 
By analyzing user response logs from our advertising system, we have found that if we group the data to several subsets according to a specific field (e.g., for a field ``user activity'', each user belongs to one of values \{`high', `moderate', `low'\}, and we can group the data to three subsets), the averaged response probability of each subset may be quite different. If the calibration omits field information, this may result in underestimation for one field value but overestimation for another one. 
Thus we focus on \textit{field-level calibration}, which learns a specific calibration function to each field value~\cite{pan2020field}. 

In this paper we propose a doubly-adaptive approach \textsf{AdaCalib} for field-level calibration.  
It learns a field-adaptive isotonic function family based on binning, where each field value is associated with a specific function with the guidance of posterior statistics. 
Besides, consider that different field values' frequencies (i.e., data sizes) may be distinct, while the reliability of posterior statistics depends on data size, a field-adaptive binning mechanism is proposed to dynamically determine the number of bins based on a field value's frequency information. 
\textsf{AdaCalib} can be integrated seamlessly into neural predictive models for online serving, compared to traditional two-stage (prediction then calibration) solution. 
The main contributions of this paper are:
\begin{itemize}
    \item  We propose \textsf{AdaCalib} for field-level calibration. It learns an isotonic function family with the guidance of posterior statistics, and can be integrated to neural predictive models. 
     
    \item  Field-adaptive mechanisms (for function learning and binning strategy) are designed to ensure that the learned calibration function is appropriate for the specific field value. 
    
    \item  Experiments on billion-size datasets verify that \textsf{AdaCalib} brings significant improvement on calibration performance. It has been deployed online and beats previous approach on core metrics. 
\end{itemize}

\section{Prerequisites}\label{method:baseline}
Calibration is usually regarded as a post-process technique. 
Given predictive models that produces un-calibrated scores, calibration aims at transforming  them to approximate posterior probabilities. We introduce representative techniques, and describe their limitations  for field-level calibration.

\textbf{Binning-based approaches}~\cite{zadrozny2001binning,zadrozny2002iso,naeini2015bbq} rank all samples based on un-calibrated scores and \textit{partition them to bins}. \textsf{Histogram Binning}~\cite{zadrozny2001binning} takes the posterior probability of a bin as the calibrated score to each sample in the bin. \textsf{Isotonic Regression}~\cite{zadrozny2002iso} adjusts bin's number and width to ensure {monotonicity}, that is, if one sample's un-calibrated score is higher than another one, its calibrated score should be also higher. This is reasonable for modern neural models~\cite{covington2016deep} because their ranking ability is usually impressive. These non-parametric approaches are effective but lack of expressiveness.  

\textbf{Scaling-based approaches}~\cite{platt1999platscaling,guo2017temperaturescaling,kull2017betacalib,kweon2022gammacalib} \textit{learn parametric functions} that map un-calibrated scores to calibrated ones. 
They assume that probabilities are approximately specific distributed, such as Gaussian distribution with same variance to positive/negative class for \textsf{Platt Scaling}~\cite{platt1999platscaling} and Gamma distribution for \textsf{Gamma Calibration}~\cite{kweon2022gammacalib}. The scaling function is defined based on the assumption and is trained with the pairs of \{un-calibrated score, binary label\}. 

\textbf{Hybrid approaches}~\cite{kumar2019hybrid,pan2020field,deng2020calibrating,huang2022mbct} borrow ideas from binning and scaling. For example, \textsf{Smoothed Isotonic Regression}~\cite{deng2020calibrating} integrates isotonic regression and linear interpolation to a unified solution. 

Most approaches only consider un-calibrated predictions as unique input for calibration, which is unaware of field information. \textsf{Neural Calibration}~\cite{pan2020field} puts forward the problem of field-level calibration by taking sample's features as the input. In this work, the field information (carried by sample's features) is used as an auxiliary module's input to interpose the final output, thus it is not directly involved in the computation of calibration function.

\section{Problem Definition}
Consider that we have trained a neural predictive model $f_\mathrm{uncalib}(\cdot)$ on a training dataset $\mathcal D_\mathrm{train}=\{(x, y)\}$, here $x$ denotes a sample's feature information (including user/item/context fields) and $y\in\{0, 1\}$ denotes a sample's binary response label: 
\begin{equation}
    \hat p=f_\mathrm{uncalib}(x)    
\end{equation}
where $\hat p\in (0, 1)$ is the prediction produced by a logistic function $\sigma(\cdot)$.  
The model is un-calibrated: its output may not reflects true likelihood because we cannot observe sample-wise likelihood. 

For a field $\mathcal Z$ having multiple possible values $\{z_1,z_2,\ldots\}$, field-level calibration is to learn a  function family $\mathcal F_\mathrm{calib}(\cdot; z)$, which post-processes the prediction $\hat p$ to calibrated probability $\hat p_{\mathrm{calib}}$:
\begin{equation}
 \hat p_{\mathrm{calib}}= \mathcal F_\mathrm{calib}\left(\hat p; z\right)\,,\text{where the field $\mathcal Z$'s value of sample $x$ is $z$}\,.   
\end{equation} 
The goal is that $\hat p_{\mathrm{calib}}$ is close to the posterior probability of samples having field value $z$. 

The metric for evaluating calibration approaches is field-level relative calibration error (Field-RCE):
\begin{equation}\small
\label{metric:calibration}
    \text{Field-RCE}=\frac{1}{|\mathcal{D}|}\sum_{z\in\mathcal Z} \frac{\left|\  \sum_{ (x^{(i)}, y^{(i)}) \in \mathcal{D}^z} \left( y^{(i)} - \hat p^{(i)}_{\mathrm{calib}}  \right)\  \right|}{ \frac{1}{|\mathcal D^z|}  \sum_{(x^{(i)}, y^{(i)})\in \mathcal{D}^z} y^{(i)}}  
\end{equation}
where $\mathcal{D}=\left\{(x^{(i)}, y^{(i)})\right\}_{i=1}^{|\mathcal D|}$ is the test dataset. The samples having field value $z$ construct a subset $\mathcal{D}^z=\left\{(x^{(i)}, y^{(i)})\mid \ z^{(i)}=z\right\}$. For the dataset $\mathcal{D}^z$, Field-RCE evaluates the deviation level of each sample's calibrated probability $\hat p^{(i)}_{\mathrm{calib}}$. 
We can see that the key of well-calibration at field-level is that, for a specific field value $z\in\mathcal Z$, the calibration function should be adaptive to $z$ for producing an appropriate calibrated probability.

\section{Proposed Approach}
We propose a doubly-adaptive approach \textsf{AdaCalib} that learns field-adaptive calibration function for each field value with the guidance of posterior statistics. 
\textsf{AdaCalib} is a hybrid approach, which takes the advantages of monotonicity and posterior integration in binning-based solutions, as well as the rich expressiveness in scaling-based solutions. 

\textsf{AdaCalib} learns a field-adaptive isotonic (monotonical) function family via binning-based posterior guidance, and each field value is associated with a specific function for calibration. 
Moreover, consider that the reliability of each bin's posterior depends on the size of samples in the bin, a field-adaptive binning mechanism is  further proposed to dynamically determine the number of bins.

\subsection{Field-Adaptive Isotonic Function Family}\label{method:function}

\subsubsection{\textbf{Binning of Samples with Un-calibrated Predictions}}
Consider that we have trained a predictive model $f_\mathrm{uncalib}(\cdot)$ that produces prediction score $\hat p\in (0,1)$ for each sample. 
For the dataset $\mathcal D_\mathrm{dev}=\{(x^{(i)}, y^{(i)})\}$ that is used for training calibration function, we employ the model $f_\mathrm{uncalib}(\cdot)$ to produce  un-calibrated predictions on each sample, which is denoted as $\hat p^{(i)}$ for the $i$-th sample. 

Given a field value $z$ of field $\mathcal Z$, to obtain the corresponding posterior statistics based on $\mathcal D_\mathrm{dev}\cup \mathcal D_\mathrm{train}$, we take all samples whose field $\mathcal Z$'s value is equal to $z$, that is, $\left\{(x^{(i)}, y^{(i)})\mid z^{(i)}=z\right\}$. 
Based on these samples' un-calibrated scores, we rank them in ascending order, and partition these samples to $K$ {equi-frequency bins}: we compute $K+1$ bound values $\left(b_1(z), b_2(z), \ldots, b_{K+1}(z)\right)$ where each bin contains same size of samples. 
For a sample $x^{(i)}$ which has field value $z$, if its un-calibrated score $\hat p^{(i)}$ meets the condition of $b_k(z) \leq \hat p^{(i)} <  b_{k+1}(z)$, we place it to the $k$-th bin.

To guide the calibration function learning, we compute {posterior statistics of each bin} based on the constructed bins.\footnote{To ensure that the averaged response rate of each bin is monotonically increasing, we can employ isotonic regression to process the original binning. However in practice we find that the final performance is not sensitive to this.} 
Then for the $k$-th bin's samples, posterior statistics include the averaged response rate and the sum of responses. We denote the posterior statistics of the $k$-th bin's samples as $\boldsymbol p_k(z)$. 

\subsubsection{\textbf{Posterior-Guided Calibration Function Learning}}
Our \textsf{AdaCalib} learns an isotonic (monotonical) function family to map un-calibrated predictions to calibrated scores. For each field value $z$, \textsf{AdaCalib} has a function specific to it for calibration.

To preserve the ranking ability of the un-calibrated model and focus more on approximating the posterior probabilities, the formulation of the calibration function is a continuous piecewise linear mapping with monotone increasing nature, where each bin has a linear mapping~\cite{deng2020calibrating,pan2020field}. 
Specifically, for each bin's samples, we adopt linear mapping to transform the un-calibrated predictions $\hat p$ to calibrated scores $\hat p_{\mathrm{calib}}$, and the key for learning calibration function is to represent the slope of each bin. 

Formally, the calibration function for $z$ is represented as:
\begin{equation}
\begin{aligned}
    \hat p_{\mathrm{calib}} &= \mathcal F_\mathrm{calib}(\hat p; z) \\
        &= \frac{\mathsf{MLP}\bigl(\boldsymbol p_{k+1}(z)\bigr)  - \mathsf{MLP}\bigl(\boldsymbol p_{k}(z)\bigr) }{b_{k+1}(z) - b_{k}(z)}\bigl( \hat p -  b_k(z) \bigr) +  \mathsf{MLP}\bigl(\boldsymbol p_{k}(z)\bigr)\,, \\  
        & \quad \mathrm{\ if\ }b_k(z) \leq \hat p< b_{k+1}(z) 
\end{aligned}
\end{equation}
where $\mathsf{MLP}(\cdot)$ denotes a fully-connected layer that enables rich expressiveness.\footnote{For $\boldsymbol p_{K+1}(z)$, we use the maximum values in $\mathcal{D}_\mathrm{train}$ as the statistics. }  
In this parameterized formulation, for samples in the $k$-th bin, the calibration function has a slope $\frac{\mathsf{MLP}\bigl(\boldsymbol p_{k+1}(z)\bigr)  - \mathsf{MLP}\bigl(\boldsymbol p_{k}(z)\bigr) }{b_{k+1} - b_{k}}$ that is determined by the posterior statistics $\boldsymbol p_{k}(z)$.\footnote{We transform posterior statistics $\boldsymbol p_{k}(z)$ to embeddings (by discretizing continuous values to IDs) as the input of  $\mathsf{MLP}(\cdot)$. }

\begin{figure}[t]
\centering
\centerline{\includegraphics[width=0.8\columnwidth]{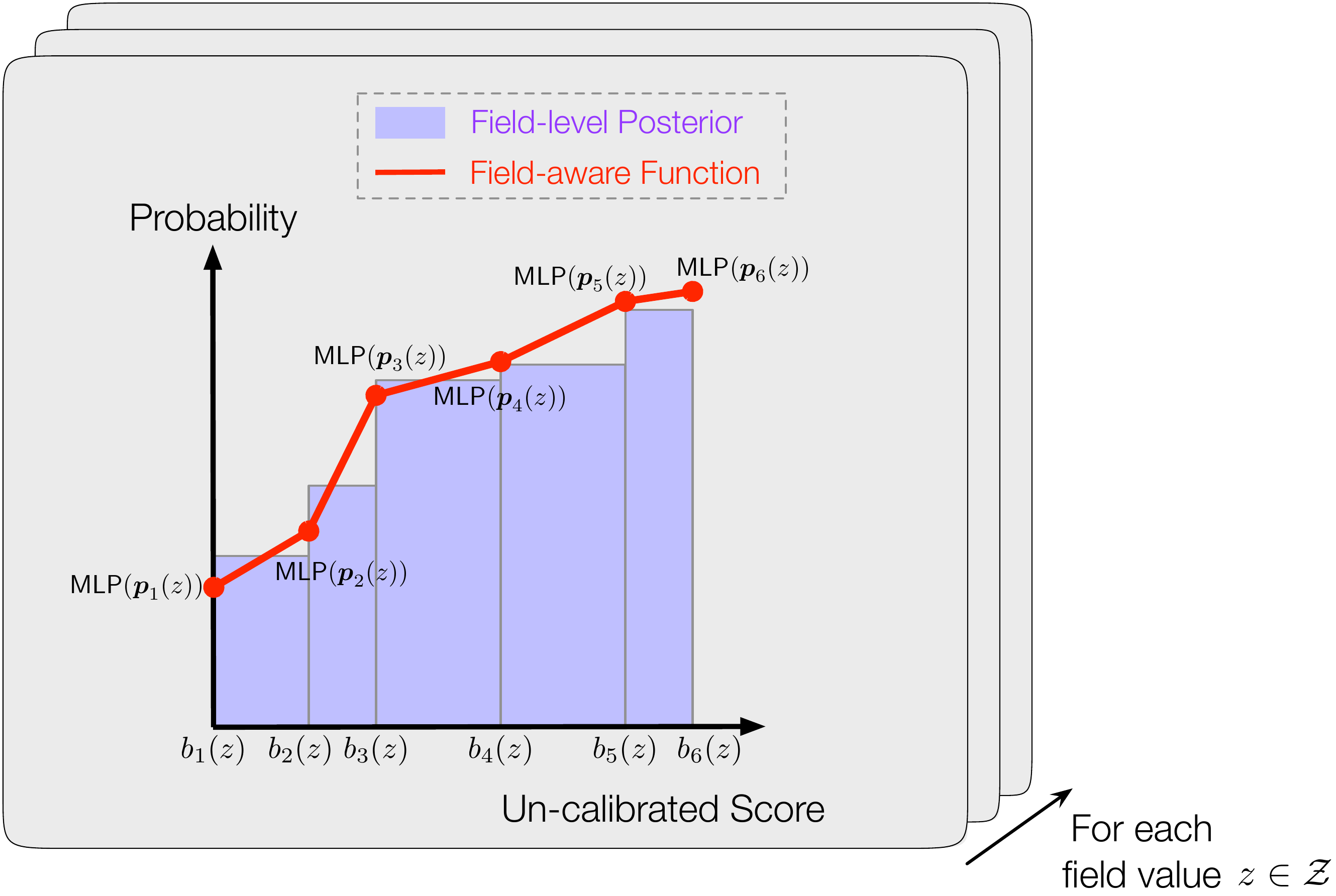}}
\caption{Learned calibration functions.}
\label{fig:ada}
\end{figure}

The posterior statistics guide the learning of calibration functions by imposing on the slope of each bin. Because the posterior statistics are computed on specific field values' bins, learned functions are adaptive to field values and perform field-level calibration. Figure~\ref{fig:ada} gives an illustration of learned calibration functions.

In practice, we perform calibration on the logit $\sigma^{-1}(\hat p)$ other than the  score $\hat p$. 
This enables the improvement of expressiveness via adding an auxiliary term $\mathsf{MLP}_\mathrm{aux}(x)$ to calibrated logit~\cite{pan2020field}, and it is possible to improve ranking ability. Then a logistic function is used to output the final calibrated score. 

\subsubsection{\textbf{Optimization}}\label{method:main_loss}
The calibration function is optimized with:
\begin{equation}\small
    \mathcal{L}(\hat p_{\mathrm{calib}}) = \ell\left(\hat p_{\mathrm{calib}}, y\right) + \sum_{k=1}^K\max\bigl\{\mathsf{MLP}\bigl(\boldsymbol p_{k}(z)\bigr)  - \mathsf{MLP}\bigl(\boldsymbol p_{k+1}(z)\bigr) , 0\bigr\}
\end{equation}
where $\ell$ is cross-entropy. The second term plays the role of a constraint, which ensures that for each bin $k$ its slope is non-negative.

\subsection{Field-Adaptive Binning}\label{method:binning}
In the above we have implemented field-adaptive calibration function, where the binning number is fixed to $K$ for all field values. 

The reliability of posterior probability depends on the size of samples in a bin. 
Different field values' frequencies may be distinct. We expect that the high-frequency field values (i.e., having sufficient samples) can be partitioned to more bins to compute posterior statistics in a fine-grained manner, while low-frequency field values is assigned with few bins to avoid inaccurate posteriors. 

We propose a field-adaptive binning mechanism to dynamically determine the number of bins. Specifically, we pre-define $n$ candidates of bin number $\{K_1, K_2, \ldots, K_n\}$, and maintain $n$ calibration function families $\left\{\mathcal F^1_\mathrm{calib}(\cdot;z), \mathcal F^2_\mathrm{calib}(\cdot;z), \ldots, \mathcal F^n_\mathrm{calib}(\cdot;z)\right\}$. 

After obtaining the calibrated scores $\left\{ \hat p^{1}_{\mathrm{calib}}, \hat p^{2}_{\mathrm{calib}},\ldots, \hat p^{n}_{\mathrm{calib}} \right\}$ produced by all calibration functions, \textsf{AdaCalib} employs hard attention to select one of them as the final calibrated probability. Formally, the attention scores $\{\alpha_1, \ldots, \alpha_n\}$ is computed by a fully-connected layer $\mathsf{MLP}_\mathrm{bin}(\cdot)$ that takes the field value $z$'s frequency statistics as input, where a gumbel-softmax operation~\cite{jang2016categorical} is used to approximate hard selection and keep the model differentiable: 
\begin{equation}\small
\begin{aligned}
    &\{\hat \alpha_{i}\}_{i=1}^n = \mathsf{MLP}_\mathrm{bin}(\boldsymbol s_z)\\
    &\alpha_{i}=\frac{\exp \left(({\log \left(\hat \alpha_{i}\right)+g_{i}})/{\tau}\right)}{\sum_{j=1}^{n} \exp \left(({\log \left(\hat \alpha_{j}\right)+g_{j}})/{\tau}\right)}, \text { where } g_{i}=-\log \left(-\log u_{i}\right), u_{i} \sim U(0,1)\\
    & \hat p_{\mathrm{calib}} = \sum_{i=1}^n  \alpha_{i}\cdot  \hat p^{i}_{\mathrm{calib}}
\end{aligned}
\end{equation}
where $\boldsymbol s_z$ denotes $z$'s frequency information including the total number of samples having value $z$ and the ID embedding of $z$. 

The optimization objective of our \textsf{AdaCalib} is: 
\begin{equation}\small
\mathcal L_\mathsf{AdaCalib}=\sum_{i=1}^n\mathcal L\left(\hat p^{i}_{\mathrm{calib}}\right)\,.    
\end{equation}

\begin{figure}[t]
\centering
\centerline{\includegraphics[width=\columnwidth]{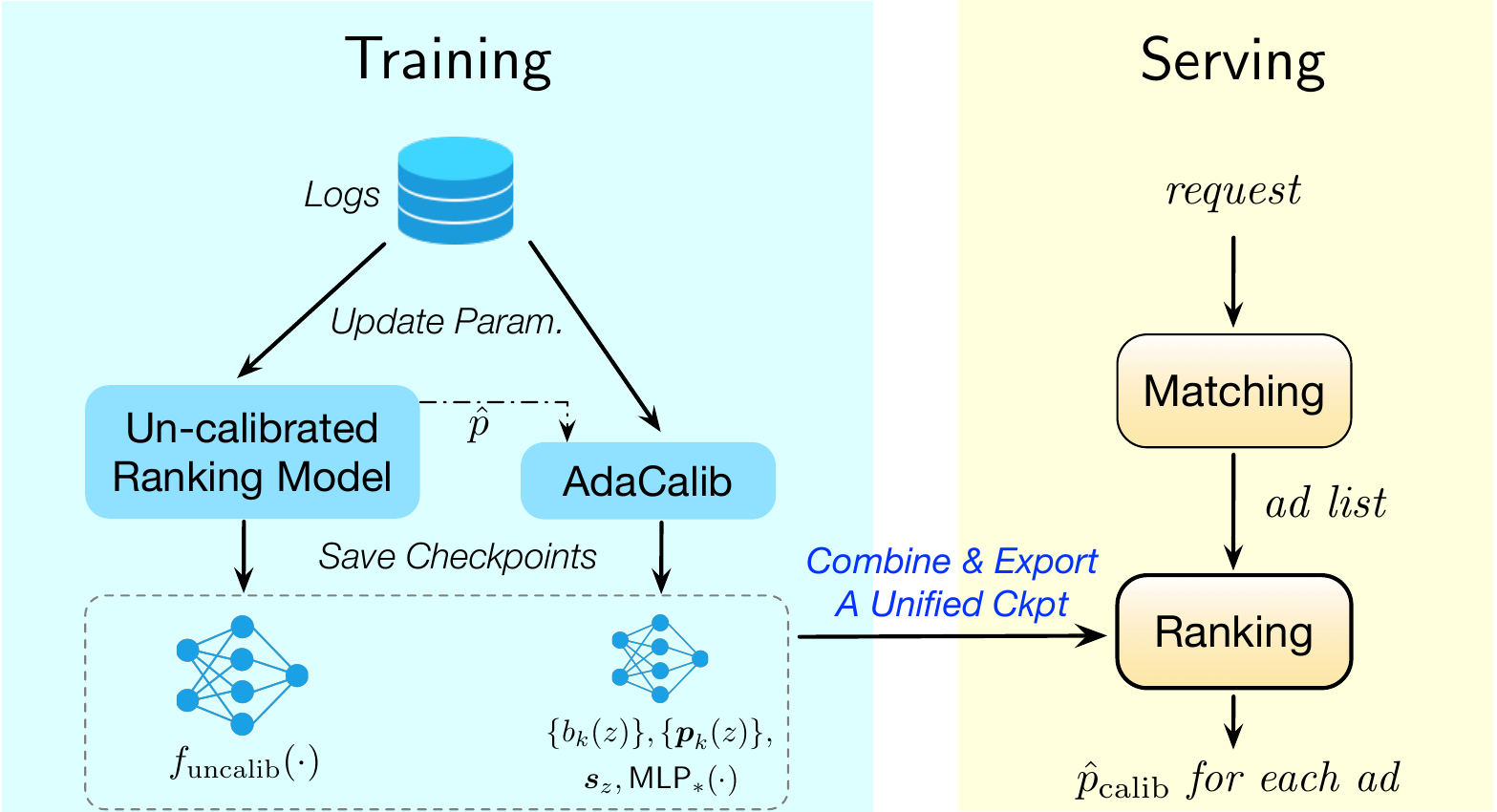}}
\caption{Training and serving of \textsf{AdaCalib}.}
\label{fig:arch}
\end{figure}

\subsection{Online Deployment}\label{method:online}
For a sample $x$ having field value $z$, \textsf{AdaCalib} takes its un-calibrated score $\hat p$, bound values of binning $\{b_k(z)\}$, posterior statistics $\boldsymbol p_{k}(z)$ and frequency information $\boldsymbol s_z$ as inputs, and the forward process is a light neural network. 
Given a neural predictive model that produces un-calibrated scores, we integrate \textsf{AdaCalib} to the predictive model via combining their checkpoints, and export a unified checkpoint for online serving. 
Figure~\ref{fig:arch} illustrates the overall procedure, and we can see that during online serving there is no need to maintain an independent calibration module for \textsf{AdaCalib}.

\section{Experiments}

\textbf{Datasets}\quad 
We conduct experiments on both public and industrial datasets. 
We collect click and conversion logs from our advertising system to construct two industrial datasets, performing calibration on CTR and CVR prediction tasks. The dataset for calibrating CTR prediction contains 1.2 billion impressions, and we choose the field ``item ID'' that has 969 unique values. We split the log by timestamp, where the first 5 days for $\mathcal D_{\mathrm{train}}$, the 6th day for $\mathcal D_{\mathrm{dev}}$ and the 7th day for $\mathcal D_{\mathrm{test}}$. The dataset for calibrating CVR prediction contains 0.2 billion clicks, and we choose the field ``item ID'' that has 542 unique values, where the first 60 days for $\mathcal D_{\mathrm{train}}$, the next 7 days for $\mathcal D_{\mathrm{dev}}$ and the last 7 days for $\mathcal D_{\mathrm{test}}$. 
For public dataset, we use the click logs~\cite{ma2018entire} to calibrate CTR prediction, which contains 80 million impressions. We choose the field ``user group ID'' that has 13 unique values. Because the provided training/test datasets do not contain timestamp information, we split the training set with the proportion 1:1 to $\mathcal D_{\mathrm{train}}$ and $\mathcal D_{\mathrm{dev}}$.

\begin{table*}[t]
\footnotesize
\caption{Results of competitors for calibrating CTR and CVR predictive models on public and industrial datasets. $\ddagger$ denotes that the performance outperforms the second-best approach at the significant level of $p<0.05$ (paired t-test).}
\centering
\begin{tabular}{clcccccccccccc}
\toprule
\multirow{2}*{\textbf{Type}}  &  \multirow{2}*{\textbf{Approach}} & \multicolumn{4}{c}{\textbf{CVR Task} (industrial data)}& \multicolumn{4}{c}{\textbf{CTR Task} (industrial data)}  & \multicolumn{4}{c}{\textbf{CTR Task} (public data)}  \\
\cmidrule(lr){3-6}\cmidrule(lr){7-10}\cmidrule(lr){11-14}
   &   & Field-RCE  & LogLoss  &  Field-AUC &  AUC & Field-RCE  & LogLoss  &  Field-AUC &  AUC  &  Field-RCE  & LogLoss  &  Field-AUC &  AUC \\   
\midrule
{No Calib.}  & N/A     & 0.3311 & 0.1460 & 0.7833  & 0.8039  & 0.2592 & 0.0819 & 0.5879 & 0.6513   &  0.3258 & 0.1681 & 0.5760 & 0.5765\\
\cmidrule(lr){1-14}
\multirow{2}*{Binning} & \textsf{HistoBin}& 0.3060 & 0.1460 & 0.7833 & 0.8039 & 0.2537 & 0.0819 & 0.5878 & 0.6513  & 0.0452 & 0.1657 & 0.5760 & 0.5765\\
& \textsf{IsoReg}& 0.2329 & 0.1455 & 0.7833 & 0.8039   & 0.2538 & 0.0819 & 0.5877 & 0.6513 & 0.0430 & 0.1645	& 0.5760 & 0.5765\\
\cmidrule(lr){1-14}
{Scaling} & \textsf{GammaCalib} & 0.2095  & 0.1451 & 0.7833 & 0.8039   & 0.2540 & 0.0819 & 0.5878 & 0.6513  &   0.0394  & \textbf{0.1630} & 0.5760 & 0.5765
\\
\cmidrule(lr){1-14}
\multirow{3}*{Hybrid} & \textsf{SIR} & 0.2579 & 0.1461 & 0.7833 & 0.8039 & 0.2532  & 0.0819 & 0.5878 & 0.6513 & 0.0428 & 0.1644 & 0.5760 & 0.5765\\
& \textsf{NeuCalib}   & 0.1920 & 0.1448 & 0.7833 & \textbf{0.8056}  & 0.1815 & \textbf{0.0814} & 0.5888 & {0.6666}   &0.0485 & 0.1646 & \textbf{0.5897}	& 0.5903 \\
& \textsf{AdaCalib}   & \textbf{0.1666}$^\ddagger$  & \textbf{0.1445} &\textbf{0.7843} & {0.8054}  & \textbf{0.1626}$^\ddagger$  &\textbf{0.0814} & \textbf{0.5899} &\textbf{ 0.6680} & \textbf{0.0147}$^\ddagger$ & 0.1653 & \textbf{0.5897} & \textbf{0.5905} \\
\bottomrule
\end{tabular}
\label{results:main}
\end{table*}

\textbf{Comparative Approaches}\quad 
We choose representative calibration approaches as competitors.  Binning-based competitors are \textsf{Histogram Binning} (\textsf{HistoBin})~\cite{zadrozny2001binning} and \textsf{Isotonic Regression} (\textsf{IsoReg})~\cite{zadrozny2002iso}.  Scaling-based competitor is \textsf{Gamma Calibration} (\textsf{GammaCalib})~\cite{kweon2022gammacalib}. Hybrid approaches are \textsf{Smoothed Isotonic Regression} (\textsf{SIR})~\cite{deng2020calibrating} and \textsf{Neural Calibration} (\textsf{NeuCalib})~\cite{pan2020field}.

We conduct CTR calibration on both datasets as well as CVR calibration on industrial datasets. Thus we trained three un-calibrated  models, and model architecture is composed of four-layered fully-connected layers~\cite{covington2016deep}. For comparative approaches that need binning, we set the number of bins to 10 empirically. 
The candidate set of bin numbers in \textsf{AdaCalib} is set to \{5, 10, 20\} for CVR task and \{2, 4, 8\} for CTR task. All calibration approaches are trained using $\mathcal D_\mathrm{dev}\cup \mathcal D_\mathrm{train}$ because in practice we found that the calibration performance is better in this way. 

\textbf{Evaluation Metrics}\quad 
We evaluate all comparative approaches from two perspectives: 1) field-level calibration performance, evaluated by the metric Field-RCE in Equation~\ref{metric:calibration}. 2) field-level ranking ability, evaluated by Field-AUC (a.k.a. weighted AUC in previous work~\cite{he2016ups}), which represents whether the calibration approach hurts the ranking of the un-calibrated model. 
Besides, we also report the overall (i.e., field-agnostic) calibration and ranking performances for reference, which are evaluated by LogLoss and AUC respectively. We do not consider expected calibration error (ECE) as an overall metric because if a model simply predicts the averaged response rate for all samples, its ECE is well but such model is non-useful.

\begin{table}[t]
\footnotesize%\scriptsize
\centering
\caption{Ablation study for \textsf{AdaCalib}.}
\begin{tabular}{cccccc}
    \toprule
    \multicolumn{4}{c}{\textbf{Variants of \textsf{AdaCalib}}} & \multicolumn{2}{c}{\textbf{Metric}} \\
    \cmidrule(lr){1-4}\cmidrule(lr){5-6}
     Posterior & Adaptive & Adaptive & Auxiliary & \multirow{2}*{Field-RCE} & \multirow{2}*{Field-AUC} \\
    Guidance  & Function & Binning & $\mathsf{MLP}_\mathrm{aux}(x)$ \\
    \midrule
    \checkmark & \checkmark & \checkmark &  \checkmark &    0.1666 & 0.7843 \\ % 
     \checkmark & \checkmark & \checkmark &  $\times$ &        0.1778 & 0.7828 \\ % 
    \checkmark & \checkmark & $\times$ &   $\times$ &         0.1946 & 0.7827 \\ % 
   \checkmark & $\times$ & $\times$ &  $\times$ &            0.2120 & 0.7828 \\ % 
     $\times$ & $\times$ & $\times$ &  $\times$ &            0.2348 & 0.7833 \\ % 
    \bottomrule
\end{tabular}
\label{results:ablation}
\end{table}

\subsection{Results and Discussion}
\subsubsection{\textbf{Main Results}}
Table~\ref{results:main} shows the results for calibrating CTR and CVR predictive models, where the first line lists the metrics of un-calibrated models. 
In terms of ranking ability, we observe that calibration usually do not hurt AUC and Field-AUC of un-calibrated models. \textsf{NeuCalib} and \textsf{AdaCalib} slightly improve AUC and Field-AUC because they employ auxiliary $\textsf{MLP}_\mathrm{aux}(x)$ that improves the expressiveness of calibration models.

In terms of the calibration performance, we can see that traditional binning- and scaling-based approaches cannot achieve large improvement on the metric Field-RCE. Thus, the approaches that do not consider field information during calibration are less effective to field-level calibration. 
The field-level competitor \textsf{NeuCalib} outperforms most approaches on Field-RCE and LogLoss, demonstrating that taking field value into account is indeed effective for calibrating model predictions at field-level. 
Our \textsf{AdaCalib} takes the advantages of binning's posterior statistics and parametric function's expressiveness to learn field-adaptive calibration function family, and achieves the state-of-the-art calibration performance for all three predictive models.

\subsubsection{\textbf{Ablation Study}}
Table~\ref{results:ablation} shows several variants of \textsf{AdaCalib} on calibrating CVR models. 
Specifically, the variant without posterior guidance means that the two slope parameters of each bin are directly set as two learnable parameters and do not consider posterior statistics $\boldsymbol{p}_k(z)$. The variant without adaptive function means that binning is performed on global dataset other than the subset specific to field value $z$. The variant without adaptive binning means that we employ a fixed binning number for all field values. As the results shown in Table~\ref{results:ablation} (note that the last two lines' variants do not perform field-level calibration), we can see that all components contribute to the calibration performance, which verifies the effectiveness of each component in \textsf{AdaCalib}.

\begin{figure}[t]
\centering
\centerline{\includegraphics[width=0.75\columnwidth]{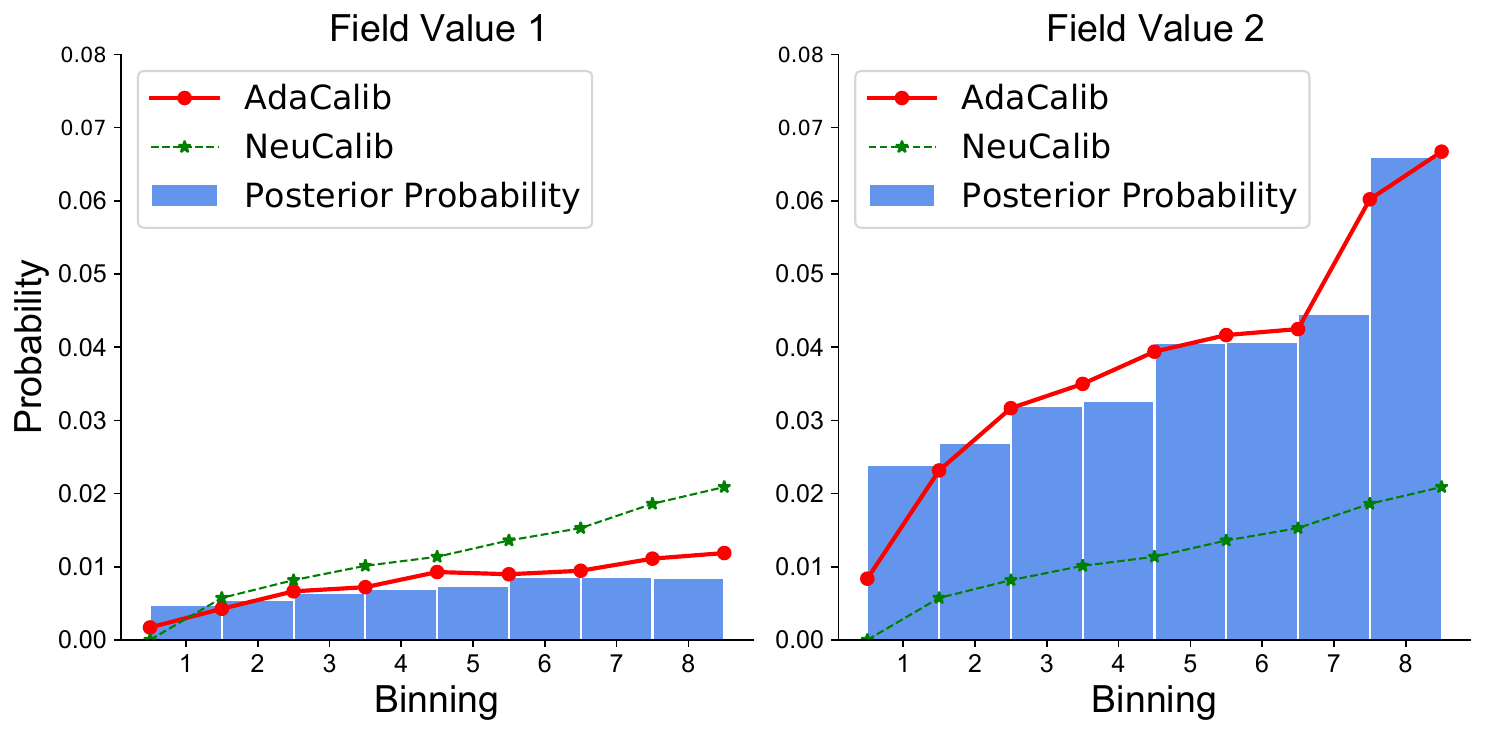}}
\vspace{-1em}
\caption{Calibration functions of \textsf{AdaCalib} and \textsf{NeuCalib}.}
\label{fig:illustration}
\end{figure}

\begin{figure}[t]
\centering
\centerline{\includegraphics[width=0.4\columnwidth]{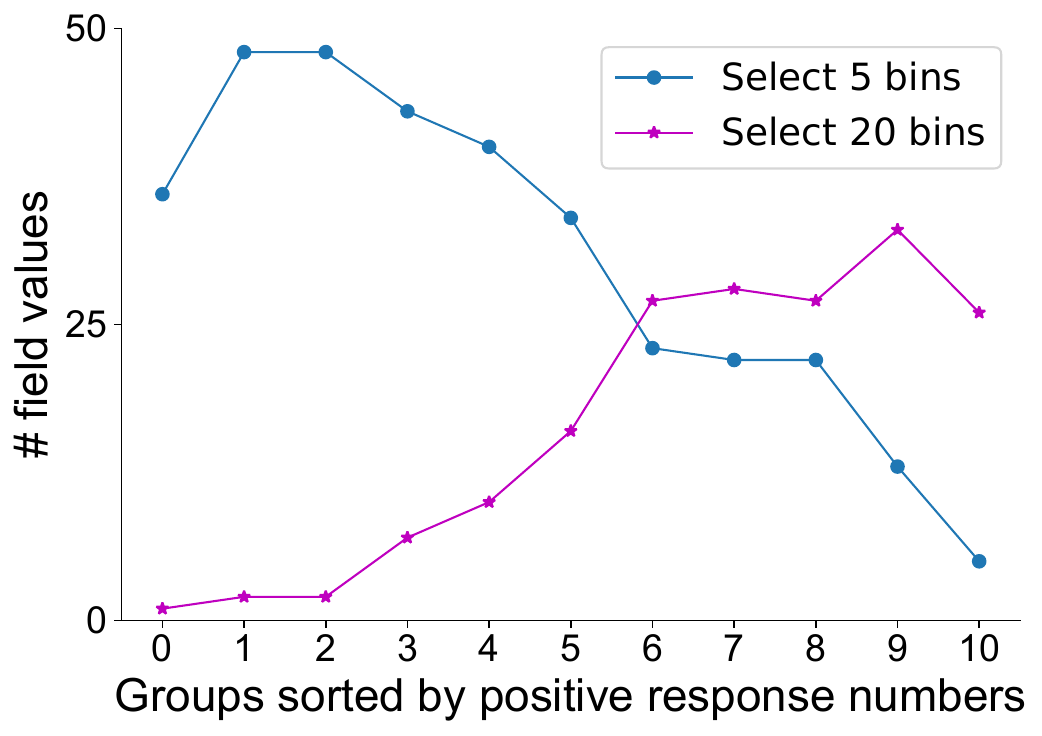}}
\vspace{-1em}
\caption{The number of field values selecting 5 or 20 bins.}
\label{fig:binning}
\end{figure}

\subsection{Further Analysis}
\subsubsection{\textbf{Comparison of Learned Functions}}
Figure~\ref{fig:illustration} shows the learned calibration functions (i.e., piece-wise linear mapping for all bins) of \textsf{AdaCalib} and \textsf{NeuCalib} for two distinct field values. Recall that we perform calibration on logits, thus for visualization we employ logistic function to transform the slope parameter $\mathsf{MLP}\bigl(\boldsymbol p_{k}(z)\bigr)$ to probability space (y-axis). Because the two approaches employ different binning strategies, to ensure that their calibration functions are comparable, the x-axis represents bin ID other than boundary value. Compared to \textsf{NeuCalib}, \textsf{AdaCalib} learns more appropriate functions that approximate field-level posterior probabilities, verifying the effect of field-adaptive mechanisms.

\subsubsection{\textbf{Effect of Field-Adaptive Binning}}
We rank all field values' subsets $\{\mathcal D^z\}$ based on their positive response numbers (which reflects the reliability of posterior statistics), and then merge all $|\mathcal Z|$ subsets to 10 groups. Let the 1st group has fewest positive responses and the 10th group has most ones. 
Figure~\ref{fig:binning} shows that in each group, the number of field values that select 5 or 20 bins via adaptive binning mechanism. 
We can see that with the increasing of response numbers, more field values in the group tends to select the large binning number (i.e., 20 bins)  other than the small numbers. This verifies that the field-adaptive binning mechanism can choose appropriate binning numbers for specific field values.

\subsection{Online A/B Test}
We conduct online experiments using our advertising system's CVR prediction module for one week. Compared to \textsf{SIR}, it brings \textbf{+5.05\%} lift on CVR and \textbf{+5.47\%} lift on Gross Merchandise Volume (GMV).

\section{Conclusion}
We propose a doubly-adaptive calibration approach \textsf{AdaCalib}. Field-adaptive mechanisms for function learning and binning strategy ensure that the learned calibration function is appropriate for specific field value. Both offline and online experiments verify that \textsf{AdaCalib} achieves significant improvement. 
In future work, we shall explore how to improve \textsf{AdaCalib} under data sparsity.

\section*{Acknowledgments}
We thank all the anonymous reviewers to their insightful comments. 

\bibliographystyle{ACM-Reference-Format}
\bibliography{src-shortversion.bib}

\end{document}